\pdfoutput=1

\documentclass[11pt]{article}

\usepackage{EMNLP2022}

\usepackage{times}
\usepackage{latexsym}
\usepackage{comment}

\usepackage[T1]{fontenc}

\usepackage[utf8]{inputenc} 

\usepackage{microtype}

\usepackage{multicol}
\usepackage{graphicx}
\usepackage{booktabs}
\usepackage{tabularx}
\usepackage{xspace}
\usepackage{times}
\usepackage{textcomp}
\usepackage{multirow}
\usepackage{amsmath}
\usepackage{graphicx}
\usepackage{multirow,bigdelim,dcolumn,booktabs}
\usepackage{tikz-dependency}
\usepackage{float}
\usepackage{caption}
\usepackage{color, colortbl}
\usepackage{amssymb}

\newcommand{\keeptag}[0]{\texttt{\textcolor{teal}{K}}}
\newcommand{\deletetag}[0]{\texttt{\textcolor{red}{D}}}

\usepackage{multirow}

\usepackage[ruled,vlined]{algorithm2e}

\newcommand{\ED}[0]{{\sc EdiT5}}
\newcommand{\FE}[0]{{\sc Felix}}
\newcommand{\LT}[0]{{\sc LaserTagger}}
\newcommand{\bv}[1]{\mathbf{#1}}

\title{EdiT5: Semi-Autoregressive Text Editing with T5 Warm-Start}

\author{
Jonathan Mallinson \\
\\
\And
Jakub Adamek \\
\hspace{5cm} Google Research \\ 
\hspace{5cm}\texttt{\{jonmall,enkait,emalmi,severyn\}@google.com}\\
\And
Eric Malmi \\ 
\\
\And
Aliaksei Severyn \\ 
}

\date{}

\begin{document}
\maketitle
\begin{abstract}
We present \ED{}\footnote{Code and pre-trained models \url{https://edit5.page.link/code}} – a novel semi-autoregressive text-editing model designed to combine the strengths of non-autoregressive text-editing and autoregressive decoding. \ED{} is faster during inference than conventional sequence-to-sequence (seq2seq) models, while being capable of modeling flexible input-output transformations.

This is achieved by decomposing the generation process into three sub-tasks: (1) \textit{tagging} to decide on the subset of input tokens to be preserved in the output, (2) \textit{re-ordering} to define their order in the output text, and (3) \textit{insertion} to infill the missing tokens that are not present in the input. The \textit{tagging} and \textit{re-ordering} steps, which are responsible for generating the largest portion of the output, are non-autoregressive, while the \textit{insertion} step uses an autoregressive decoder. 

Depending on the task, \ED{} on average requires significantly fewer autoregressive steps, demonstrating speedups of up to 25x when compared to seq2seq models. Quality-wise, \ED{} is initialized with a pre-trained T5 checkpoint yielding comparable performance to T5 in high-resource settings  when evaluated on three NLG tasks: Sentence Fusion, Grammatical Error Correction, and Decontextualization while clearly outperforming T5 in low-resource settings.
\end{abstract}

\section{Introduction}

\begin{figure}[tb]
\centering
\includegraphics[scale=0.6]{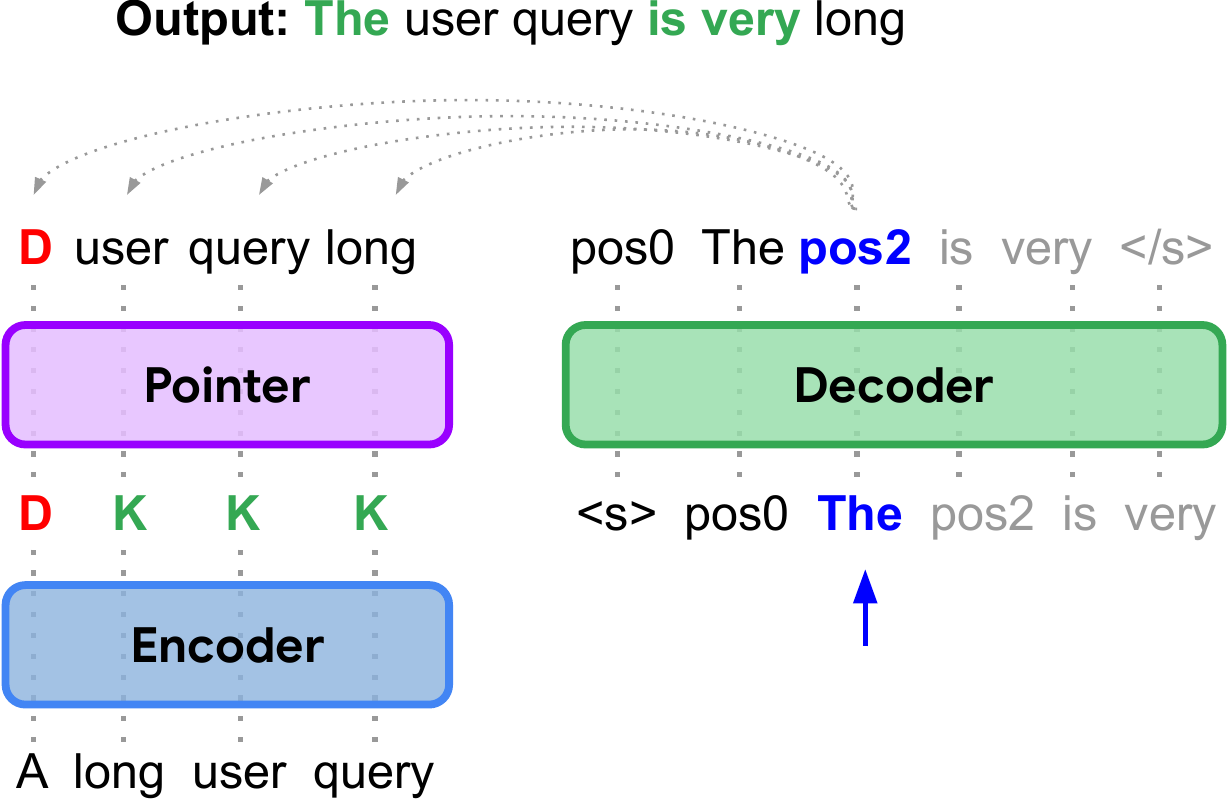}
\caption{EdiT5 transforms the input text \textit{A long user query} into the output \textit{The user query is very long} by first generating a sequence of edit tags \deletetag{} \keeptag{} \keeptag{} \keeptag{} (where \keeptag{} stands for keeping and \deletetag{} for deleting the input token), re-ordering the input tokens with the pointer network, and infilling missing tokens into the source sequence with an autoregressive decoder which jointly predicts the text spans (\texttt{The} and \texttt{is very}) and the position where to insert them (\texttt{pos0} and \texttt{pos2}). The blue arrow shows how the token \texttt{pos2} is predicted conditioned on the prefix \texttt{<s> pos0 The} generated thus far. The dotted arrow lines depict the encoder-decoder cross attention over the re-ordered input tokens and edit tags. }
\label{fig:edit5}
\end{figure}

Pre-trained seq2seq models such as T5 \cite{t5}, BART \cite{lewis2019bart}, and MASS \cite{song2019mass} have established strong baselines for the majority of text-to-text transduction tasks. A recent trend to massively scale up model sizes, e.g., all the way up to 540B params~\cite{chowdhery2022palm}, as well as the sizes of pretraining corpora, has further pushed the state-of-the-art without signs of reaching a plateau. From a practical point of view, running inference with such models is prohibitively expensive for most applications, which motivates the work on finding efficient recipes for model distillation, e.g., \cite{kim2016sequence} and choosing a model architecture that can provide a better trade-off between performance on a given task and inference speed. A typical choice is to distill a large language model into a smaller seq2seq model, e.g., Transformer~\cite{vaswani2017attention}. In this paper we propose a novel model architecture \ED which blends ideas from a seq2seq T5~\cite{t5} and text-editing to provide faster inference without sacrificing on task performance.

Seq2seq-based models output text token-by-token from scratch, allowing them to model any kind of input-output relationship. However, for many real-world tasks this degree of generality is unnecessary, especially for monolingual tasks where the input and output texts have relatively high degrees of overlap. In such cases  a natural approach is to cast conditional text generation as a text-editing task, where the model learns to construct target texts by applying a set of edit operations to the inputs~\cite{malmi2022text}. Typically the set of edit operations is defined ahead of time \cite{omelianchuk2020gector,LaserTagger,awasthi2019parallel}, which on the one hand limits the flexibility of the model to reconstruct arbitrary output texts from the inputs, but on the other, leads to  latency improvements as the limited set of allowed operations significantly reduces the output vocabulary of the decoder. In this paper, we propose an approach which is both fast at inference time and flexible, able to model arbitrary rewrites. 

\paragraph{Faster inference.}
A common method for achieving low latency in serving models is to reduce their size, thus reducing their computational cost. Doing so naively, however, often leads to inferior model quality, and much work has gone into finding better methods for model size reduction, such as distillation \cite{kim2016sequence}.

Regardless of model size, one of the major contributors to the total inference time for seq2seq models is the decoder, which generates the output sequence step-by-step. \ED{} also relies on an autoregressive decoder, but generates the majority of the output sequence with its tagging and pointing networks, and as such the decoder makes far fewer steps. 

\paragraph{Flexible text-editing.}
Recent text-editing approaches, e.g.,~\cite{awasthi2019parallel,LaserTagger}, are not as powerful as general purpose seq2seq approaches when it comes to modeling arbitrary input-output text transductions. \ED{} supports open-vocabulary generation by relying on an autoregressive decoder. In the extreme case, where there is no overlap between the source and the target texts, it reduces to a vanilla seq2seq model generating the entire output from scratch. However, when the input and output  overlap, it can benefit from the \textit{tagging} and \textit{pointer} networks to reconstruct the bulk of the output text that is further infilled (refined) by the autoregressive decoder. 

\paragraph{Warm start.}
Training a high-precision text generation model typically requires large amounts of high-quality supervised data. Self-supervised techniques based on text in-filling~\cite{Bert2Bert, BART, t5} have been shown to provide a crucial advantage over non-pre-trained models especially in low-resource settings. Hence, we design \ED{} to be able to benefit from already existing pre-trained language models (specifically T5), where the final model is directly fine-tuned on the downstream task.  \\

\ED{} decomposes the generation task into three steps: \textit{tagging}, \textit{pointing} and \textit{insertion}
(see Fig.~\ref{fig:edit5}). The tagger and pointer networks decide which source tokens to preserve and in which order they should appear in the output, thus allowing for arbitrary word dropping and reordering. The tagger is implemented using a  non-autoregressive feedforward network, and pointing is implemented using a novel non-autoregressive pointing mechanism \citep{PointerNetworks} combined with sinkhorn layers~\citep{mena2018learning}. The insertion network inserts/infills words which are present in the target sequence but do not appear in the source sequence. The network is implemented using an autoregressive transformer decoder, which attends to the tagged, reordered source sequence. The decoder predicts both the locations of where the token spans should be infilled, as well as the spans themselves.

We evaluate \ED{} on three distinct text generation tasks: Sentence Fusion, Grammatical Error Correction (GEC), and Decontextualization, comparing to recent text-editing approaches and T5. Each task is unique in the editing operations required and the amount of training data available, which helps to better quantify the value of modeling decisions we have integrated into \ED{}.

Additionally, we explore the impact of training data size and model size on \ED{}. Finally we quantify the latency of \ED{}, providing a detailed analysis and comparison to T5.

\section{Model description}

The model architecture of \ED{} resembles a vanilla Transformer~\cite{vaswani2017attention} composed of an \textbf{encoder} and a \textbf{decoder}. \ED{} decomposes the generation of a text $\bv{y}$ from an input $\bv{x}$ into three parts: predicting a sequence of edit tags $\bv{y}^t$ (indicating whether a token from the input should be copied to the output), a permutation of the input tokens $\bv{\pi}$ (indicating the order that copied tokens should appear in in the output), and a sequence of tokens $\bv{y}^d$ (indicating additional tokens that should be in the output, and where in the permuted input they should be inserted). $\bv{y}^t$ and $\bv{\pi}$ are modeled by the \textbf{encoder}, and $\bv{y}^d$ by the \textbf{decoder}.

There are multiple ways to choose the triple ($\bv{y}^t$, $\bv{\pi}$, $\bv{y}^d$) for a given ($\bv{x}$, $\bv{y}$) pair. During dataset creation we choose a single such triple for each training pair (see section \ref{sec:dataset-contruction} for details), in which case the probability of $\bv{y}$ can be expressed as:

\begin{align}
 P(\bv{y}|\bv{x}) := & \left( \prod^{|\bv{y}^d|}_i  P(\bv{y}^d_i| \bv{y}^{d}_{<i}, \bv{y}^t, \bv{\pi}, \bv{x}) \right) \nonumber \\
 & * P(\bv{\pi}| \bv{y}^t, \bv{x}) * P(\bv{y}^t| \bv{x}) 
\end{align}

During inference, we first greedily set $\bv{y}^t$ to maximize the third term, then $\bv{\pi}$ to maximize the second term, and finally $\bv{y}^d$ to maximize the first term. The output text $\bv{y}$ is realized by applying the tags $\bv{y}^t$ and permutation $\bv{\pi}$ to the input sequence $\bv{x}$ and then inserting the tokens $\bv{y}^d$.

\subsection{Text-editing encoder}

The \ED{} encoder consists  of three steps:
 encoding, tagging, and  pointing.

\paragraph{Encoder.}
The source sentence ${\bv{x}}$  is first encoded using $N$ transformer layers into the hidden representations $\bv{h}$.

\paragraph{Tagging.}
The tag sequence $\bv{y}^t$ is constructed as follows: source tokens that must be copied are assigned the KEEP tag, tokens not present in the output are marked by the DELETE tag.
Tags are predicted by applying a single transformer layer followed by a classification layer to the output of the encoder $\mathbf{h}$, which is trained using cross-entropy:

\begin{equation}
   \mathcal{L}_{tagging} = - \sum_{j}^{|\bv{x}|} \log P ({y}^t_j| f_t(\bv{h})_j)
\end{equation}
 where $\bv{y}^t$ are the gold tags,  $j$ is the index of the source token, and $f_t$ is a transformer layer followed by a classification layer. During inference we use \emph{argmax} to determine the tags, whereas during training we use the gold tags. The encoder hidden state is then updated to take these tags into account:
 
 \begin{equation}
     \bv{h}^{t}_j = f_{te}([\bv{h}_j; TE(\bv{y}^t_j)])
 \end{equation}
Where $TE$ is a tag embedding layer, whose output is concatenated to the original hidden representation of the source sequence, before a feed-forward layer $f_{te}$ is applied. 

\paragraph{Pointing.} 
In many tasks it is helpful for the model to be able to rearrange the kept input tokens. For example, we can grammatically correct the sentence \emph{Who you are?} to \emph{Who are you?} purely by reordering tokens from the input. In \ED{} this is made possible thanks to its pointing mechanism. In contrast, in text editing approaches such as~\citet{LaserTagger,EditNTS}, correcting this sentence involves first deleting the words \emph{you are} and then recreating them in the right order.

\begin{figure}[tb]
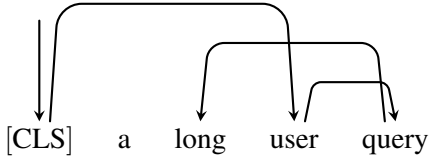

\centering
\begin{dependency}[hide label]
\begin{deptext}[column sep=.4cm, row sep=.0ex]

$[$CLS$]$ \& a \& long \& user \& query \\
\end{deptext}
\deproot[edge style=thick, edge unit distance=2.5ex]{1}{root}{}
\depedge[edge style=thick]{1}{4}{}
\depedge[edge style=thick]{4}{5}{}
\depedge[edge style=thick]{5}{3}{}
\end{dependency}

\caption{Pointing mechanism to transform 
\textit{``a long user query"} into \textit{``user query long"}.}
\label{fig:pointing_example}

\end{figure}

Given a sequence $\bv{x}$ and the predicted tags $\bv{y}^t$, the re-ordering model generates a permutation $\bv{\pi}$. Our implementation is based on a pointer network \citep{PointerNetworks}, where an attention mechanism points to the next token. We follow \citet{mallinson-etal-2020-felix} which, unlike previous approaches where a decoder state attends over an encoder sequence, applies intra-attention, where source tokens attend to all other source tokens. As such the output of this model is a series of predicted pointers, where each source token predicts the token that comes after it. $\bv{\pi}$ can easily be constructed by daisy-chaining these predicted pointers together, as seen in Fig.~\ref{fig:pointing_example}. 
 We calculate attention using key-query attention, where we include an additional transformer layer prior to the key network:

\begin{equation}
    \alpha_{m,j} = f^{q}(\bv{h}^{t})_m  \boldsymbol{\cdot} f^{k}(\bv{h}^{t})_j
\end{equation}

Where $\alpha_{m,j}$ is the unnormalized attention, $f^{q}$ is the query network, a single feed-forward layer, and $f^{k}$ is the key network, a transformer layer followed by a single feedfoward layer. 

Unlike \citet{mallinson-etal-2020-felix}, we ensure a valid permutation is formed, i.e. no token is pointed to twice, by using sinkhorn layers \citep{mena2018learning}, which normalizes over both the rows and the columns of the intra-pointer attention $\alpha$. Sinkhorn layers are defined as: 

\begin{align}
S^0 &= \exp(\alpha) \\
S^i &= T_c(T_r(S^{i-1}(\alpha))
\end{align}
where $T^{j,m}_c (X) = \frac{X_{j,m}}{\sum_l X_{l,m}}$ is the column normalization operator and $T^{j,m}_r (X) = \frac{X_{j,m}}{\sum_l X_{j,l}}$ 
is the row normalization operator.

The loss for the pointing network is defined as: 

\begin{equation}
   \mathcal{L}_{pointing} =  CE(\pi|S(\alpha))
   \end{equation}
Where CE is the cross-entropy loss. During inference we use argmax to determine $\pi$.

We use additional positional embeddings to update the hidden states with their new position (offset from 0). For example if \emph{Who you are?} was reordered into \emph{Who are you?}, the position information would be updated as \textsubscript{0}{Who} \textsubscript{2}{you} \textsubscript{1}{are} \textsubscript{3}{?}.
\begin{equation}
   \bv{h}^p_j = (\bv{h}^t_j + \bv{PE}(\pi_j))
\end{equation}
where $PE$ are learnt absolute positional embeddings \cite{BERT}. These additional positional embeddings are masked out for those source words which do not appear in the target sequence.  Finally we apply a transformer encoder layer to $\bv{h}^p$ forming the final encoded representation of the sequence $\bv{h}^f$. $\bv{h}^f$ captures the edits as well as the original sequence $\bv{x}$, and the decoder attends to this representation.

\paragraph{Decoder.}
We use a standard transformer decoder, which is tasked with inserting tokens which are in the output sequence but don't appear within the input sequence. \ED{} takes advantage of the pre-training of a T5 model, where T5 was pre-trained to infill missing spans. When pre-training T5 uses special tokens \emph{\textlangle{}pos\_i\textrangle{}} to indicate where missing spans should be inserted, as demonstrated in Figure 3. \ED{} re-purposes these special tokens, using them to indicate at which position new tokens should be infilled. I.e. \emph{\textlangle{}pos\_1\textrangle{}}, indicates that the tokens should be inserted after the first token.  As such the decoder first decodes a special position token and then decodes the inserted tokens which should appear after this token. For example to insert \emph{the cat} after the first token, the decoder generates: \emph{\textlangle{}pos\_1\textrangle{} the cat}. The decoder is trained with a standard cross-entropy loss:

\begin{equation}
   \mathcal{L}_{insertion} = - \sum_{i}^{|\bv{y}^d|} \log P (y^d_i| \bv{y}^{d}_{<i}, h^f)
\end{equation}
Where $i$ is the decoder index, and ${h}^f$ is the encoder output. The loss for the entire model is defined as the sum of the three individual losses:
\begin{equation}
\mathcal{L} = \lambda_{1}\mathcal{L}_{tagging}   + \lambda_{2}\mathcal{L}_{pointing}    + \lambda_{3}\mathcal{L}_{insertion}    
\end{equation}
where $\lambda_1$, $\lambda_2$ and $\lambda_3$ are hyper-parameters determining the relative importance of tagging, pointing and insertion losses in the final loss.

\paragraph{Pre-training.}
\label{sec:pre-train}

\begin{figure}
\footnotesize
{\footnotesize
\setlength{\parindent}{3pt}
\begin{tabularx}{\columnwidth}{lX}
\textbf{Source/Target:} & \texttt{a long user query .} \\[0.4em]

\textbf{T5 Input:} & \texttt{a} [X]  \texttt{user query} [Y] \\[0.4em]
\textbf{T5 decoder:} &[X] \texttt{long} [Y]  \texttt{.} \\[0.4em]
\textbf{EdiT5 Input:} & \texttt{user a query the}
\\[0.4em]
\textbf{EdiT5 tagger:} & \texttt{\keeptag{}\hspace{0.6cm} \keeptag{} \keeptag{} \hspace{0.6cm} \deletetag{}}  \\[0.4em]
\textbf{EdiT5 pointer:} & \texttt{a user query}  \\[0.4em]
\textbf{EdiT5 decoder:} & [0] \texttt{long} [2] \texttt{.} \\[0.4em]
\end{tabularx}

}
    \caption{Example pre-training noise for T5 and \ED{}. \keeptag{} and \deletetag{} indicate keep and delete tags resspectivly, and [0] indicates \emph{pos0}.  }
    \label{fig:noise}
\end{figure}

While we initialize \ED{} from T5 base, T5 was pre-trained with 12 decoder layers, and for \ED{} we use a single decoder layer. To account for this change in the decoder layers, we perform additional pre-training. We use a pre-training objective which combines a T5 style span insertion task, with a generic text-editing denoising task, as used in BART~\cite{BART}. A source sentence is corrupted by dropping, swapping and adding spans (an example can be seen in Figure~\ref{fig:noise}), and we task our model to reconstruct the original sentence. By introducing noise we are able to train the tagger to detect incorrect spans, and the pointer to reorder the sentence. The decoder then behaves like the T5 pre-training objective inserting the content of missing spans. Unlike BART's pre-training, our approach is computationally cheap, as we do not decode the entire sequence when training, instead just decoding the missing spans.  

\paragraph{Dataset construction.} \label{sec:dataset-contruction} When constructing the training dataset, there are many possible combinations of $\bv{y}^t$, $\bv{\pi}$ and $\bv{y}^d$ which could produce $\bv{y}$. For instance, all source tokens could be deleted and the decoder could then produce all the target tokens. However to minimize latency, we wish to make the number of inserted tokens (i.e. the number of decoder steps) as small as possible, and maximize the number of kept tokens.

To produce alignments from a target sequence to a source sequence, we iterate left-to-right through characters in the target sequence, trying to find spans of target characters which appear in the sequence of source tokens, as described in Algorithm \ref{alg} (see Appendix \ref{sec:alignment-algorithm}). Each source token can only be aligned to a single target span. Those target spans that can't be aligned are instead inserted after the closest previous aligned source token. In cases where there are multiple possible alignments, e.g. the same token appears multiple times in the source, we align the target character span to produce the longest contiguous span of source tokens aligned with the target, i.e. where source tokens appear one-after-another in the target sequence. To find the longest contiguous span we compare the contiguous overlap between source and target for each possible alignment.

\section{Experiments}
We evaluate \ED{} on three distinct text-editing tasks: Sentence Fusion, Grammatical Error Correction, and Decontextualization. In addition to reporting previously published results for each task, we also compare to \FE{}~\cite{mallinson-etal-2020-felix}, a recent non-autoregressive text-editing model, and a strong pre-trained T5 baseline implemented in the T5X framework \cite{roberts2022t5x}.

\paragraph{Modeling.}
For \ED{} we initialize with a T5 base model with a 12-layer Transformer encoder, and single-layer Transformer decoder. Our code is based on the Tensorflow Model Garden's~\cite{tensorflowmodelgarden2020} TF2 version of T5. After initializing with the T5 checkpoint, we further pre-train on the denoising objective (see Section~\ref{sec:pre-train}) using the C4 corpus~\cite{t5}, training for 100k steps.

For all experiments  \ED{} is trained using AdamW \cite{loshchilov2017decoupled}, additionally the   learning rate was decayed using the validation set, and exact match is used for checkpoint selection. Tokenization is  based on T5's  SentencePiece vocabulary \cite{kudo2018sentencepiece}, with a vocabulary size of 32k. We, however, modify the vocabulary, removing tokens which have punctuation as a suffix, and replacing them with additional span insertion special token, giving \ED{} 512  span insertion special token. Unless otherwise stated, we use an input sequence length of 128. We performed minimal hyper-parameter selection, which is discussed in the Appendix.

\paragraph*{Task Analysis.} 
The chosen tasks cover a diverse set of edit operations and a wide range of dataset sizes, varying from under 11 thousand data points to over 4.5 million. Table \ref{tbl:ter} provides dataset statistics including: the size, input sequence length, output sequence length for seq2seq models, the output sequence length for \ED{}, and the translation error rate (TER) \cite{snover2006study}  between the source and target sentences. We use TER to highlight unique properties of each task.  

From Table \ref{tbl:ter} we see that for all tasks  \ED{} requires significantly fewer decoder steps than a seq2seq model, which results in significant latency savings. We also see that decontextualization has the longest input and output sequences, where the maximum input length of decontextualization is 512 tokens. Decontextualization  has the highest TER, with the major contribution being deletion, which is due to the input sequence consisting of a paragraph, whereas the output is a single sentence. In contrast GEC, has the shortest input and output sequence, with the majority of the dataset consisting of a single input and a single output sentence. GEC has the lowest TER, however it has the highest insertion TER. Sentence fusion consists of two sentences being rewritten into a single sentence, and has a middling TER and sequence lengths. It also has the fewest substitutions.

\begin{table*}[t]
\centering
\begin{small}
\begin{tabular}{lrrrrrrrrrr}
\toprule
Dataset & Size & $L_{\text{src}}$ & $L_{\text{tgt}}$ & {\sc E5}-Ins & TER & Ins & Del & Sub & Shft  \\ 
\midrule
Sentence fusion &  4.5M & 42.5 & 41.1 & 5.8 &  10.92 & 2.49& 04.91& 3.75& 0.62 \\  
GEC & 2.3M & 24.3 & 24.7 & 4.6 & 09.72 & 2.99 & 01.19& 5.05 & 0.49 \\
Decontextualization & 11K & 193.9 & 49.1 & 7.2 & 84.80 & 0.28 &90.64 &6.43 & 2.65&    \\
\bottomrule
\end{tabular}

\caption{Statistics across tasks: size of the dataset (Size), source length in tokens ($L_{\text{src}}$), target length in tokens ($L_{\text{tgt}}$), EdiT5 insertion tokens ({\sc E5}-Ins), and TER scores, including number of insertions (Ins), deletions (Del), substitutions (Sub), and shifts (Shft). Token counts are measured using a sentencepiece tokenizer and averaged over the development set. }
\label{tbl:ter}

\end{small}

\end{table*}

\subsection{Sentence Fusion}
Sentence Fusion is the task of fusing independent sentences into a coherent output sentence(s) \cite{geva2019discofuse}. It requires operations such as inferring the appropriate discourse connective, pronominalization, reordering the text to introduce  relative clauses, and changing the order of the input sentences.

\paragraph{Data.} We use the “balanced Wikipedia” portion of the DiscoFuse dataset~\cite{geva2019discofuse} and also study the impact of training data size by creating four  additional smaller subsets of DiscoFuse consisting of: 450,000 (10\%), 45,000 (1\%), 4,500 (0.1\%) and 450 (0.01\%) data points.

\paragraph{Setup.} Following \citet{geva2019discofuse}, we report \textit{Exact match}, which is the percentage of exactly correctly predicted fusions. In addition to the T5 baseline and the text-editing baselines \LT{}~\cite{LaserTagger}, \FE{}~\cite{mallinson-etal-2020-felix}, and Seq2Edits~\cite{stahlberg2020seq2edits}, an autoregressive text-editing model, we also report state-of-the-art seq2seq models ROBERTASHARE~\cite{rothe-etal-2020-leveraging}, based on ROBERTA large, and AugBERT \cite{ben-david-etal-2020-semantically}, based on BERT base. Additionally, we measure the impact of our pre-training  (Section 2.1) initializing \ED{} with a T5 checkpoint, without additional pre-training.

\paragraph{Results.}

\begin{table}[tb]
\centering
\resizebox{\columnwidth}{!}{%
\footnotesize
\begin{tabular}{p{2.cm}p{0.75cm}p{0.55cm}p{0.55cm}p{0.55cm}p{0.55cm}p{0.55cm}p{0.75cm}}
\toprule
 & \#Params & 100\% & 10\% & 1\% & 0.1\% & 0.01\% & latency \\ 
\midrule
\LT{} & {110M} & 53.80 & 47.31 &  38.46 & 25.74 & 12.32 & -\\
\FE{} & 220M & 61.31 & 52.85 & 45.45 & 36.87 &16.96 & \textbf{1.8} \\
Seq2Edits & 279M & 61.71 & - & - &-  &- &- \\
\ED{} & 141M & 64.95 & 59.26 & \textbf{52.09} & \textbf{43.83}  & \textbf{28.64} &{2.2} \\ 
\quad - pre-training & 141M & 65.16 & 59.27 & 50.39 & 34.18 & 1.90 & {2.2} \\
\midrule
T5 base & 220M & 65.52 & \textbf{59.75} & 50.75 & 33.84 & 10.75 & 52.7 \\
ROBERTA & 380M & \textbf{66.6} & - & - & - & - & - \\
AugBERT & 157M & 65.0 & - & - & - & - & -\\
\bottomrule
\end{tabular}
}

\caption{Sentence fusion results (Exact Match, lower-cased) under various data conditions,  latency (ms), and number of parameters.}
\label{tbl:df}

\end{table}

From the top section in Table \ref{tbl:df} we first observe that \ED{} strongly outperforms other text-editing methods. Next it performs comparably to T5 in high-resource settings (100\% and 10\%), where it's just 0.5 points lower in exact match than T5, whilst achieving a latency that is 25 times faster, and using fewer parameters. The current SOTA, ROBERTASHARE, which outperforms \ED{} by 1.5 points, is based on the ROBERTA large checkpoint which overall has more parameters and a larger encoder. In low-resource settings, \ED{} clearly outperforms T5 by up to 18 points (0.01\%, i.e. 450 training examples).

The results in Table \ref{tbl:df} additionally demonstrate that the significant improvements of \ED{} over Felix in high/medium-resource settings do not stem from \ED{} pre-training. With 450 datapoints, pre-training is critical since there’s a larger mismatch between \ED{} and T5 checkpoints than there is between Felix and BERT checkpoints. We additionally ablated the impact of sinkhorn layers, and found that under the 100\% data condition there was a modest decrease in performance (0.5 exact match points).

\subsection{Decontextualization}
Sentence decontextualization task was introduced by~\citet{choi2021decontextualization}. The goal is to rewrite an input sentence to make it stand-alone without the original context.

\paragraph{Data.}
We use the train, dev and test data from \citet{choi2021decontextualization}, where sentences were selected from Wikipedia passages. Human annotators were asked to rewrite them, if possible, to be interpretable and grammatical without the context. We compare against T5 base, T5 xxl, \FE{}, and a copy baseline. All models use a sequence length of 512. 

\paragraph{Metrics.} 
Following \citet{choi2021decontextualization}, we report exact match, exact match when a sentence needs to be rewritten and SARI F1 (deletion and addition) on unigrams~\cite{xu2016optimizing}. 

\paragraph{Analysis.} Results in Table \ref{tbl:decon} show that \ED{} achieves a higher exact match scores, and SARI delete score when compared to T5 base, with a significant drop in latency and using fewer parameters. T5 base achieves significantly higher SARI add, suggesting its better at inserting new tokens, which is unsurprising as \ED{} is primarily focused on copying the source sequence. Both T5 and \ED{} achieve significantly higher numbers than \FE{}. \ED{} and T5 base, however, still achieve a significantly lower score than the T5 xxl, which can be explained by the difference in model size.

\begin{table}[tb]
\centering
\resizebox{\columnwidth}{!}{%
\footnotesize
\begin{tabular}{p{1.6cm}p{0.8cm}p{0.8cm} p{0.7cm}p{0.8cm}p{0.8cm}l}
    \toprule
    & \#Params  & EM & EMc & ADD & DEL & latency \\  
    \midrule
    Repeat & -& 36 & 0 & 0 &0  & -\\
     T5 xxl & 11B  & \textbf{52} & \textbf{32} & \textbf{43} & \textbf{47}  & - \\ 
    \midrule
    \FE{} & 220M & 32 & 10 & 28 & 32 & 4\\
    \ED{} & {141M} & {48} & {23} & 31 & {41}  &\textbf{3.8}\\ \midrule
       T5 base* &  220M & 40& 21 &{36} & 40 & 75\\
\bottomrule
\end{tabular}
}

\caption{Decontextualization results, including exact match (\emph{EM}, exact match on those sentences which need rewriting {\emph{EMc}}, SARI \emph{ADD}, SARI \emph{DEL}ete, latency (ms), and number of parameters. * indicates scores were calculated by running the models provided by \citet{choi2021decontextualization} on the test set.} 
\label{tbl:decon}

\end{table}

\subsection{Grammatical Error Correction}
GEC requires systems to identify and fix grammatical errors in a given input text.

\paragraph{Data.}
We evaluate on the standard GEC test set BEA~\citep{bryant-etal-2019-bea}, and  use BEA-DEV for checkpoint selection. For pre-training we use an artificial GEC dataset C4\_200M of 200M sentences~\cite{stahlberg2021synthetic}. We then fine-tune on  cLang-8~\cite{rothe2021a}, a distilled version of the Lang-8 learners corpus \cite{mizumoto2011mining}.

\paragraph{Setup.} 
We report \emph{ERRANT} F0.5 scores for BEA. We report additional gT5/gFelix baseline numbers from \citet{rothe-etal-2020-leveraging}, where T5/Felix models were trained only on cLang-8. For pre-training we sampled 0.2\% examples from the training set to use as a development set, and train till convergence as measured on this development set.

We additionally measure the impact that model size has on quality and latency, training T5 and \ED{} small, base, and large models. To make the latency comparison fairer, we also train single-decoder-layer variants of the T5 models we call T5 Slim. To further ensure a fair latency comparison between \ED{} and T5 we use the same framework for both models. Additionally, we do not perform \ED{} specific pre-training.

\paragraph{Results.}

From Table \ref{tbl:gec}, we see that all models outperform their equivalent gT5/gFelix models, which is not surprising as the latter models were trained on less data. A surprising result is that the T5 slim variants achieve comparable scores to the full T5 models while having significantly lower latency. Comparing \ED{} against T5 models, we see up to $\sim$1 point differences in F0.5 scores between models of the same size (small/base/large), however \ED{} produces speed ups between 10x and 25x. 

\begin{figure}[tb]
    \centering
    \includegraphics[width=\columnwidth]{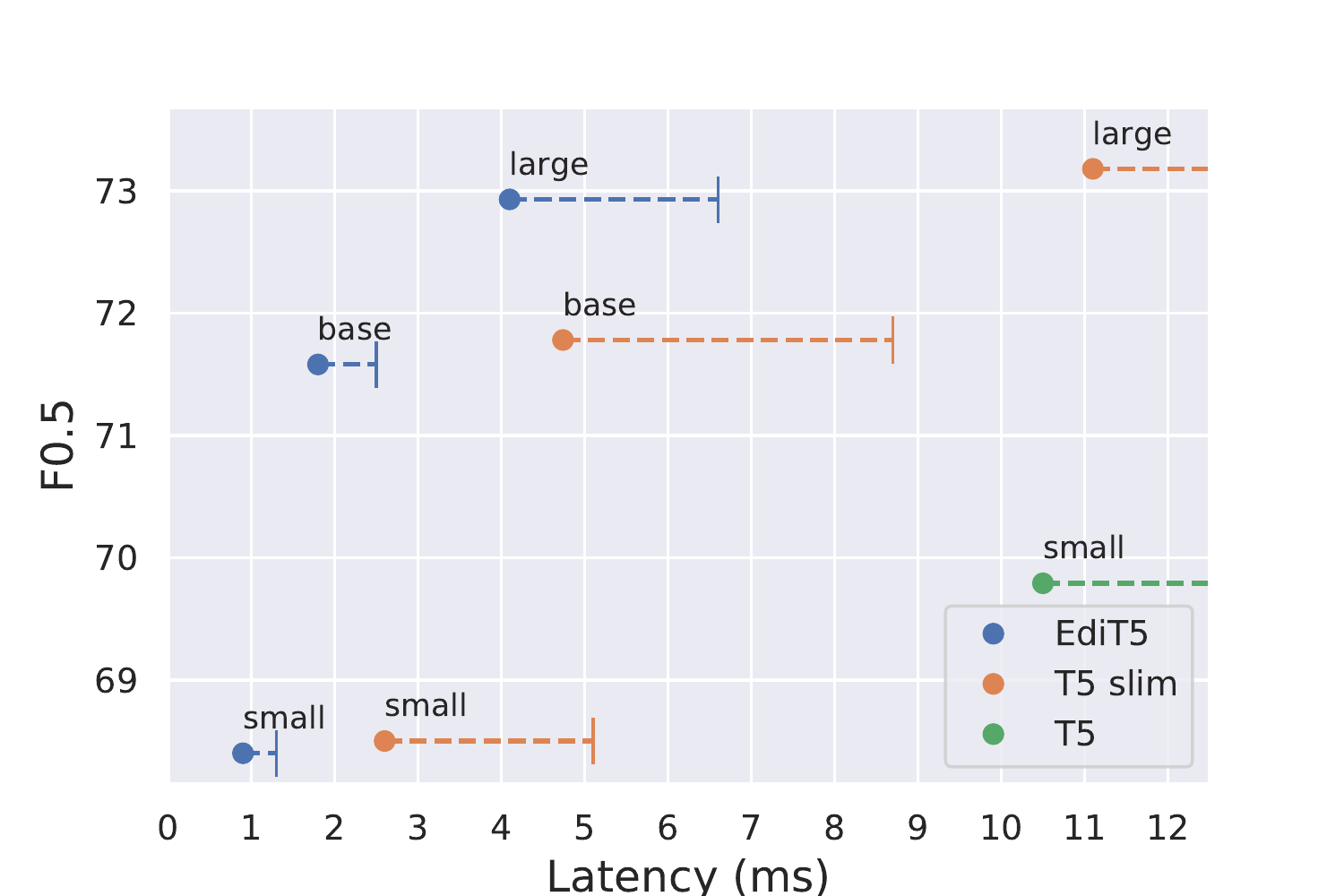}
    \caption{Mean and 95\% percentile latency for T5, T5 slim and \ED{} across model sizes on BEA. }
    \label{fig:latency}

\end{figure}

In Figure \ref{fig:latency}, we study the latency--quality trade-offs of T5, T5 slim, and \ED{} models. We omit Felix from this analysis, because Felix achieves a significantly lower score.  We focus on the 95 percentile latency, as it is often the case that users require that a model returns a result within a fixed latency budget.  We see that \ED{} drops less than 0.25 F0.5 points comparing across model sizes, whilst being significantly faster. Additionally for a given latency budget of 5ms, no full T5 model would fit, and only the T5 slim small would fit, whereas both \ED{} small and base fit. Comparing \ED{} base against T5 slim small, we see that \ED{} scores  3 F0.5 points higher, whilst being faster. For any latency budget under 20ms, \ED{} is quicker and offer better results than T5 and T5 slim. For latency budgets above 20ms, T5 slim large scores slightly (<0.25 F0.5) higher than \ED{}, and if latency is not a factor then gT5 xxl should be used.


\begin{table}[tb]
\footnotesize
\resizebox{\columnwidth}{!}{%
\begin{tabular}{lllllll}
\toprule
Model                & \#Params & F0.5  & Mean & Median     & 95\%    & Speed Up  \\    \midrule
gT5 small       &   76M    & 65.01 &      -        & -       &  -     &   -        \\
gT5 base        &      248M   & 69.39 &         -       &    -  &    -   &  -         \\
gT5 large       &      783M  & 72.06 &         -        &   -   &   -    &    -       \\    
gT5 xxl & 11B &  \textbf{75.88}      &     -     &   -   &    -  &    -          \\ \midrule
gFelix base & 220M & 59.05 &  -     &   -   &    -  &    -  \\ \midrule
T5 small        &   76M     & 69.79  & 10.5          & 9.2  & 21.0  & 3.5x     \\
T5 base         &    248M    &  72.39 & 35.5         & 31.2  & 74.1  & 1.0x       \\
T5 large        &   783M     & 73.43 & 92.4         & 81.3  & 184.8 & 0.4x   \\   \midrule
T5 slim small   &   55M     & 68.50 & 2.6          & 2.3    & 5.1    & 14.5x   \\
T5 slim base    &      144M  & 71.78 & 4.7          & 4.3  & 8.7  & 8.5x   \\
T5 slim large   &       391M & 73.18 & 11.1         & 10.1 & 20.0  & 3.7x   \\    \midrule
Felix base & 220M & 63.50& 1.8& 1.8& 1.8 & 41.2x\\
\midrule
\ED{} small         &    50M      & 68.40  &\textbf{0.9}          &\textbf{0.8}   &\textbf{1.3}   &\textbf{57.0x} \\
\ED{} base          &  141M      & 71.58 & 1.8          & 1.6    & 2.5   & 29.6x \\
\ED{} large          &    391M     & 72.93 & 4.1          & 3.9    & 6.6   & 11.2x    \\ \bottomrule
\end{tabular}
}
\caption{GEC F0.5 results for gT5, gFelix, T5, T5 slim, Felix, and \ED{}; number of parameters; mean, mode and 95 percentile latencies (in milliseconds); we also present  \textit{speed up}, the ratio of 95 percentile latency to T5 base.}
\label{tbl:gec}
\vspace{-0.1cm}
\end{table}

\section{Latency analysis}

The tasks on which \ED{} outperforms seq2seq models in latency are those that have  overlap between sources and targets, but it's unclear  how much overlap is required for \ED{} to produce latency savings. To answer this question, we split \ED{} base, T5 base and T5 slim base into components whose latencies we measure separately and compare. Details on how latencies are measured can be found in the Appendix \ref{sec:latency-measurement}.

A seq2seq model decomposes into two parts: the encoder (we include the input embedding here, so we refer to this as encoder* below), and the decoder. \ED{} has both of these parts, but also includes a third part (which we call its overhead), comprising of pointer realization and additional transformer layers. To make our analysis simpler and more task-agnostic, we make two simplifying assumptions. First, we assume the worst-case that no tokens are deleted by \ED{} and there are no padding tokens in the input\footnote{The pointer realization  runs for exactly input-length steps.}, in practice this is not the case, and provides significant latency savings for \ED{}. Second, we assume that decoder latency is linear in the number of decoder steps\footnote{This ignores decoder self-attention, but is justified when the number of decoder steps is small.}. Both of these assumptions benefit the latency of seq2seq models more than \ED{}.
 
\paragraph{Results.}

In Table~\ref{tbl:components} we present latencies of encoder*, worst-case \ED{} overhead and the per-step latency of a decoder under various input-length conditions. We see the overhead added by \ED{} even in the worst-case is small.

From these results we can derive a simple rule for when \ED{} will provide a net latency benefit. Compared to T5 slim base\footnote{The overhead is smaller than 1 step of T5 base.}, \ED{} base must save on average 4 decoder steps with an input length of 128, and 7 steps with an input length of 512.

Finally, collating the results in Table~\ref{tbl:components} with the number of decoder steps performed by \ED{} and T5 in Table~\ref{tbl:ter}, we see that whereas in T5 the decoder latency dominates the latency of encoder*, in \ED{} this is no longer the case. For instance for GEC, at 24.7 decoder steps on average required to construct the output, T5 slim spends 3.7x more time in its decoder than in encoder*. \ED{} however spends less time in its decoder than in encoder*, as such the encoder* is now the latency bottleneck.

\begin{table}[]
\centering
\footnotesize
\begin{tabular}{p{4cm}p{1.2cm}p{1.2cm}} \toprule
Component & Len. 128 & Len. 512 \\ \midrule
Encoder* & 0.98 & 2.65 \\
Worst-case \ED{} overhead & 0.49 & 1.16 \\
1 layer decoder per-step & 0.15 & 0.17 \\
12 layer decoder per-step & 1.26 & 1.47 \\  \bottomrule
\end{tabular}
\caption{Mean latencies (in milliseconds, $\pm$ 0.01ms) measured for the components of \ED{} and T5 models for various input lengths. \ED{} overhead is normally input dependent, but  we estimate worst-case latency.}
\label{tbl:components}
\end{table}

\section{Related work}

T5 \cite{t5} is a pre-trained, Transformer-based \cite{vaswani2017attention} encoder-decoder model which has become a general-purpose tool for a variety of sequence-transduction tasks, establishing many new state-of-the-art results \cite{t5,rothe2021a}. However, two considerable challenges hindering the productionizing of T5-based models are the high latency caused by autoregressive decoding and the need for having a relatively large number of training examples despite the fact that pre-training makes T5 more sample efficient. Recently, it has been found that the sample efficiency problem can be mitigated by performing in-context few-shot learning, but this typically requires scaling up the model size even further \cite{brown2020language,chowdhery2022palm}, increasing the latency.

To reduce latency, a number of non-autoregressive (NAT) seq2seq methods have been proposed for neural machine translation~\cite{gu2018nonautoregressive,LevenshteinTransformer,du2021order} but a quality gap compared to autoregressive methods still exists. To decrease the gap, it is common to run the NAT methods iteratively, which, however, limits the inference speed advantage over autoregressive methods \cite{lee2018deterministic}. In contrast, we show that for tasks where inputs and outputs overlap, we can maintain an order-of-magnitude speed-up without compromising on the model quality by treating the problem as a text-editing task and producing the output in a single pass.

A number of text-editing models have been proposed as a faster and more sample efficient alternative to seq2seq models like T5 \cite{awasthi2019parallel,LaserTagger,omelianchuk2020gector,mallinson-etal-2020-felix}. Another recently proposed approach to speed up the inference time of Transformer models is called \textit{aggressive decoding}~\cite{sun2021instantaneous,ge2022lossless}.

Closest to our work, \citet{mallinson-etal-2020-felix} show that adding pointing mechanism for reordering and a separate insertion model allow their text-editing model, \FE{}, to produce an arbitrary output in a flexible manner. \FE{} is a non-autoregressive model which first predicts the tokens to keep, their order, and the locations at which to insert new tokens. 
Then it runs a separate model based on a BERT masked language model for inserting new tokens. In contrast, \ED{} employs a single, end-to-end model which has an autoregressive insertion component. This enables more accurate insertions, while keeping the latency low, given that most of the tokens can be copied from the source non-autoregressively. Other text-editing models that employ autoregressive insertion include EditNTS~\cite{EditNTS}, the text-normalization model by \citet{zhang2019neural}, Seq2Edits~\cite{stahlberg2020seq2edits}, ESC~\cite{chen2020improving} and LEWIS~\cite{reid-zhong-2021-lewis}. However, unlike \ED{}, these models perform also the edit operation prediction autoregressively, making them potentially slower at inference time.

\section{Conclusions}
In this paper we have proposed \ED{} a low latency solution to text generation, that achieves comparable or betters results, across three distinct tasks, to a strong T5 baseline whilst achieving inference latencies that are up to 25x quicker than the baseline model.  

In the future we wish to explore the following ideas: 1) The impact of distillation for \ED{}.  Distillation has previously been shown to be particularly advantageous to non-autoregressive models. 2) Exploring the impact that quantization has on both latency and quality.  3) Applying \ED{} to additional languages. \ED{} makes no language specific assumptions and we plan to apply it to languages other than English. 

\section*{Limitations}

A limitation of \ED{}, and text-editing models in general, is the assumption of overlapping text between the input and output sequences. For instance, in machine translation the overlap between source and target is minimal to none. As such \ED{} would decode the entire target sequence, thus offering no latency saving. 

An additional limitation is that all of our experiments were done on English tasks.
It is unclear how \ED{}'s pointing mechanism would behave with languages which have a less strict word-order, such as Czech. 

Finally, we have measured latency only on V4 TPUs, and thus it is unclear how the performance would behave on different graphics cards or on CPUs. As such to determine if \ED{} offers a good trade-off between quality and latency, one must measure latency on the target device.

\section*{Acknowledgement}

We thank  Sebastian Krause, Sascha Rothe, and Hongkun Yu for useful
discussions, suggestions and feedback. We also thank Shankar Kumar and Felix Stahlberg for providing feedback on an earlier draft of the paper.

\bibliographystyle{acl_natbib}
\bibliography{anthology,custom}

\appendix

\section{Alignment Algorithm}
\label{sec:alignment-algorithm}

\begin{algorithm}[]
\footnotesize
  \caption{\ED{} Alignment}
  \label{alg}
  
  \KwData{source \tcp*[l]{List of tokens}} 
\KwData{target \tcp*[l]{List of characters}}
\KwResult{alignments}
  buffer $\gets \emptyset$
  
   alignments $\gets$ []
  
   $i \gets 0 $
   
   \While{$i <$  len(target)}{
     max\_length $\gets 0$
     
      max\_index  $\gets 0$
      
     $j \gets i+1 $
    
 \While{$j <$  len(target)}{
   
    source\_index, overlap\_length  $\gets$ contiguous\_length(target[i:j],source)
   
   \If{overlap\_length $>$ max\_length}{
     
      max\_length $\gets$  overlap\_length
      
     max\_index  $\gets$ source\_index
   
  }
       $j \gets j+1 $
  }
  \eIf{max\_overlap\_length $>$ 0}{

     source[max\_index]  $\gets \emptyset$ 
     
    alignment $\gets$  (i,j,max\_index,buffer)
    
   alignments.append(alignment)
      
       buffer $\gets \emptyset$ 
       
   $i \gets  j+1$
  }{
  buffer $\gets$ buffer + target[i]
  
   $i \gets  i+1$
  }}

  \end{algorithm}

\section{Training Details} 

All models were trained on 4x4 or 8x8 TPUs, all \ED{} models completed training (including \ED{} pre-training) in under a day. T5 large pre-training large took 2 days to complete and was done using a 4x4 TPU.  

\subsection{Hyper-Parameters Selection}

For T5 we compared the  T5 1.0 and T5 1.1 version using the base model on the validation sets and found that T5 1.1 performed better, as such used T5 1.1. For \ED{} we used the BEA dev set, finding that T5 1.0 base performed better than T5 1.1 and selected 1.0 for all experiments. 

For T5 we used the recommend fine-tuning settings,  including using the adafactor  optimizer \cite{shazeer2018adafactor}, with a learning rate of 0.001. For \ED{} we used AdamW with default settings and the default learning rate of 3e-4.

\paragraph{DiscoFuse.} For both \ED{} and T5 we experimented with 3 different batch sizes 128, 256, 1024. For 100\% and 10\%, there was not a noticeable difference in the DEV set exact match performance, so we chose 1024 as it converged the quickest. For 1\% and lower, we found that a batch size of 128 performed the best on the dev set. 

\paragraph{Decontextualization.} For \ED{} we experimented with the batch size 128, 256, 1024 and found that 256 offered the best exact match and used this. We also slightly modified the pre-processing code, bracketing the target sequence with [CLS] and [SEP], which helped the alignment code.

\paragraph{GEC.} For both \ED{} and T5 we used the T5 recommended number of tokens per batch of: batch size = 512, maximum sequence length = 128. We however note that T5 used the inverse: batch size = 128, maximum sequence length = 512. For T5 and \ED{} we disabled learning rate warmup when fine-tuning on cLang-8. Two additional hyperparameters were set for \ED{}, during pre-training on C4\_200M, we noted that \ED{} train set performance was lower than T5, as such we disabled dropout on the additional \ED{} specific transformer layers. We additionally used the dev set to set the values of lambda for equation 10. We experimented with tagging/pointing $\lambda$ being 1, 2, 10, or  equal to the number of tokens. Where $\lambda$ equal to the number of tokens produced the best results. 

\section{Latency measurement}
\label{sec:latency-measurement}
To report latency for a model, we run inference on a Cloud TPU V4 chip with batch size 1 and report the time spent in computations on the device. This approach ignores some practical contributors to latency, such as memory transfers between the host and device, but we found it also reduced noise significantly, while focusing on the main performance differences between \ED{}, T5 and T5 slim (the amount of computation they each perform). To further minimize spurious latency differences, both \ED{} and the baseline models are based on the same T5 implementation, found in TensorFlow Model Garden~\cite{tensorflowmodelgarden2020}.

\end{document}